\documentclass{llncs}
\usepackage{graphicx} % Required for inserting images
\usepackage{amsmath,amssymb,amsfonts}

\usepackage{multirow}
\usepackage{hyperref}
\usepackage{booktabs}
\usepackage[table]{xcolor}
\usepackage[dvipsnames]{xcolor}  % for ForestGreen
\definecolor{ForestGreen}{RGB}{34,139,34}
\date{}

\title{\textbf{SED:Lightweight Saliency prediction for Event-based data via Distillation}}
\author{
Romaric Mazna\inst{1}
\and
Jean Martinet\inst{1}
\and
Michele Magno\inst{2}
}

\institute{
i3S/CNRS, Université Côte d'Azur
\and
ETH Zürich
}

\begin{document}

\maketitle

\begin{abstract}
Event-based saliency prediction has gained attention recently, as combining
event cameras with saliency estimation can act as an upstream stage that
naturally improves the efficiency of downstream event-based perception at the
edge. However, current approaches are either neuromorphic, underperforming on
event-based saliency benchmarks, or too heavy for resource-constrained edge
applications due to their reliance on transformers or 3D convolutions.
Drawing inspiration from efficient convolutional modules, SED and aiming to exploit
the temporal information in event data, we propose a lightweight network,
trained through knowledge distillation, built on a Depthwise Spatio-Temporal
Block (DSTconv) -- a factorization of the 3D depthwise-separable convolution.
Relative to its teacher, our model reduces the model size from $180$\,MB to
$0.32$\,MB ($562\times$) and the parameter count from $45$M to $81$k
($554\times$), while matching or outperforming it on the N-DHF1K and N-UCF
Sports datasets.. Moreover, it
generalizes strongly beyond its training distribution, transferring from
synthetic to real event data where a model trained from scratch fails.
\end{abstract}

\section{Introduction}
Human visual attention refers to the process of selectively processing areas within a visual scene based on their saliency or their relevance with respect to a goal or a task \cite{carrasco2011visual}. This selective mechanism is one of the main components underlying the efficiency of the human brain \cite{ittiComputationalModellingVisual2001,carrasco2011visual}. In computer vision, modelling this process has a long history, from early bottom-up approaches such as Itti et al. \cite{itti1998model,itti2001computational,harelGraphBasedVisualSaliency2006} to deep learning methods \cite{pan2017salgan,pan2016shallow,moradi2024salfom}. Saliency models have proven useful across many tasks, including object detection \cite{venkatesh2019saliency}, drone navigation \cite{zhang2010novel}, and video surveillance \cite{yubing2011spatiotemporal}, often improving both computational efficiency and task performance \cite{venkatesh2019saliency,shu2015joint,zhang2010novel}. These methods, however, have been developed mainly for RGB images and videos recorded at a fixed frame rate, which induces substantial redundancy in the captured data.

Event cameras offer a fundamentally different sensing paradigm. Inspired by the biological retina, they detect and record per-pixel brightness changes, providing high temporal resolution and low latency (microsecond scale), low power consumption, and high dynamic range \cite{gallegoEventBasedVisionSurvey2022}. Their bioinspired, sparse nature makes them a natural fit for attention modelling and for deployment on resource-constrained robotic platforms such as the iCub \cite{parmiggiani2012design,dangeloEventDrivenBioinspired2022} or battery-powered edge devices. Yet existing approaches to event-based attention have been either SNN-based \cite{dangeloEventDrivenBioinspired2022,gruelNeuromorphicEventBasedSpatiotemporal2022} or hand-crafted \cite{chane2024event}, which limits their performance in challenging conditions such as fast-moving objects and sensor noise. Deep learning was recently introduced to this domain by SEST \cite{mazna2026exploringdeeplearningeventbased}, which reaches state-of-the-art accuracy but at a high cost: 45M parameters and a 180\,MB memory footprint, far beyond the budget of the very devices where event cameras are most useful.

The platforms where event cameras excel -- robots, drones, wearable devices -- operate in open-world conditions, encountering scenes, lighting, and motion patterns far removed from any training set. A model that is accurate only on its training distribution is of little use in deployment. A practical event-based saliency model must therefore be not only lightweight enough to run on-device, but also robust enough to generalize to the unseen scenes it will inevitably face. These two requirements
motivate the central questions of this work: 1) Can the computational cost of deep event-based saliency models be drastically reduced without sacrificing accuracy? 2) Can such compact models be made to generalize across synthetic and real event data, where small models trained conventionally tend to fail?

We answer both affirmatively. Prior work in the RGB domain \cite{li2019spatiotemporal} suggested that model and data redundancy are the main bottlenecks to efficiency, and that compact models with reduced
input resolution can rival larger ones. We extend this line of work to the event domain and show that a lightweight convolutional model can match and even outperform a transformer teacher on in-domain event-based saliency. Trained from scratch, however, such a model overfits and fails to generalize across datasets. Our key finding is that knowledge distillation closes this gap: rather than acting merely as a compression tool, it serves as a regularizer that transfers the teacher's cross-domain robustness to a student a fraction of its size.

Our contributions are as follows:
\begin{itemize}
    \item We propose a lightweight model for event-based saliency prediction, built on
our newly introduced Depthwise Spatio-Temporal Block (DSTconv), with only
$81$k parameters and a $0.32$\,MB memory footprint -- making it, to our
knowledge, the most compact saliency model to date in both the RGB and
event-based domains.
    \item Through extensive experiments, we show that, despite its size, our
    model matches or outperforms its much larger teacher in-domain on the
    N-DHF1K and N-UCF Sports datasets.
    \item We show that knowledge distillation acts as a regularizer in this setting: it enables the compact student to generalize to
    unseen synthetic data and to real event recordings, where an identical
    model trained from scratch fails.
\end{itemize}

The paper is organized as follows. Section~\ref{sec:related_work} reviews the
literature on event-based saliency prediction and knowledge distillation.
Section~\ref{sec:approach} describes the proposed approach, including the
event representation, the DSTconv module, the network architecture, and the
distillation pipeline. Section~\ref{sec:exps} presents the experimental
setup: datasets, evaluation metrics, and implementation details.
Section~\ref{sec:results} reports both saliency and computational
performance. Finally, Section~\ref{sec:concl} concludes the paper.

\section{Related work}
\label{sec:related_work}
\paragraph{Saliency prediction in event data.}
While saliency prediction in the RGB domain is well established, with
high-performing models, large-scale datasets, and benchmarks, saliency
research in the event domain is still at an early stage. Early event-based
saliency models built on the Gestalt theory of object surroundedness, using
proto-objects and Von Mises filters through a standard CNN, and later a
spiking neural network on SpiNNaker hardware
\cite{ghosh2022event,dangeloEventDrivenBioinspired2022,iacono2019proto,gruel2025performance}.
Gruel et al. adopted adaptive mechanisms with spiking neural networks to
detect high-activity regions as salient \cite{gruelNeuromorphicEventBasedSpatiotemporal2022}.
However, these models performed poorly relative to RGB benchmarks, and each
was evaluated differently owing to the lack of saliency datasets. Recently,
\cite{chane2024event} introduced an algorithm to compute a saliency score
for each event as it occurs, together with the first large-scale event
saliency dataset providing event recordings with corresponding human
fixations; this dataset, however, is limited in motion and scene variety.
Later, \cite{mazna2026exploringdeeplearningeventbased} introduced the first
deep learning approach, based on a transformer architecture, achieving
state-of-the-art performance and narrowing the gap between the event and RGB
domains. It leverages a pretrained Swin transformer as backbone, extracting
rich hierarchical features from the four backbone stages, projecting them
through a Conv3D, and building on top a lightweight decoder of two
successive Conv3D layers that refine the features and predict the saliency
maps. The authors also introduced two synthetic datasets derived from
large-scale RGB saliency datasets to address the data scarcity, and
demonstrated the generalizability of their model on the real event dataset
of \cite{chane2024event}. Despite achieving state-of-the-art performance,
their model is not suited to edge applications due to its large size
(180\,MB). In this work, we seek a trade-off between latency and performance
by proposing an ultra-efficient network that achieves state-of-the-art
results.

\paragraph{Knowledge distillation for dynamic saliency prediction in RGB videos.}
Knowledge distillation (KD) is a well-known and widely used compression
technique \cite{bucila2006model,hinton2015distilling}, transferring
knowledge from a larger teacher to a smaller student to improve the
performance of compact models. In saliency prediction, it has been the main
route to obtaining compact models. In \cite{li2019spatiotemporal}, the
authors distilled knowledge from two teachers (spatial and temporal) into
two corresponding students, which then transferred their knowledge to a
spatio-temporal target model. \cite{fu2020ultrafast} employed a similar
framework, distilling into a student from spatial and temporal teachers,
with the aim of facilitating the rapid integration of spatio-temporal
features without loss of accuracy. More recently, Moradi et al.
\cite{moradi2026knowledge} addressed video saliency prediction by distilling
knowledge from a video foundation model into a small state-of-the-art video saliency prediction
model, demonstrating consistent gains across different benchmarks.

\section{Proposed approach}
\label{sec:approach}

\begin{figure*}[t]
    \centering
    \includegraphics[width=\textwidth, height=0.6\textheight,
        keepaspectratio]{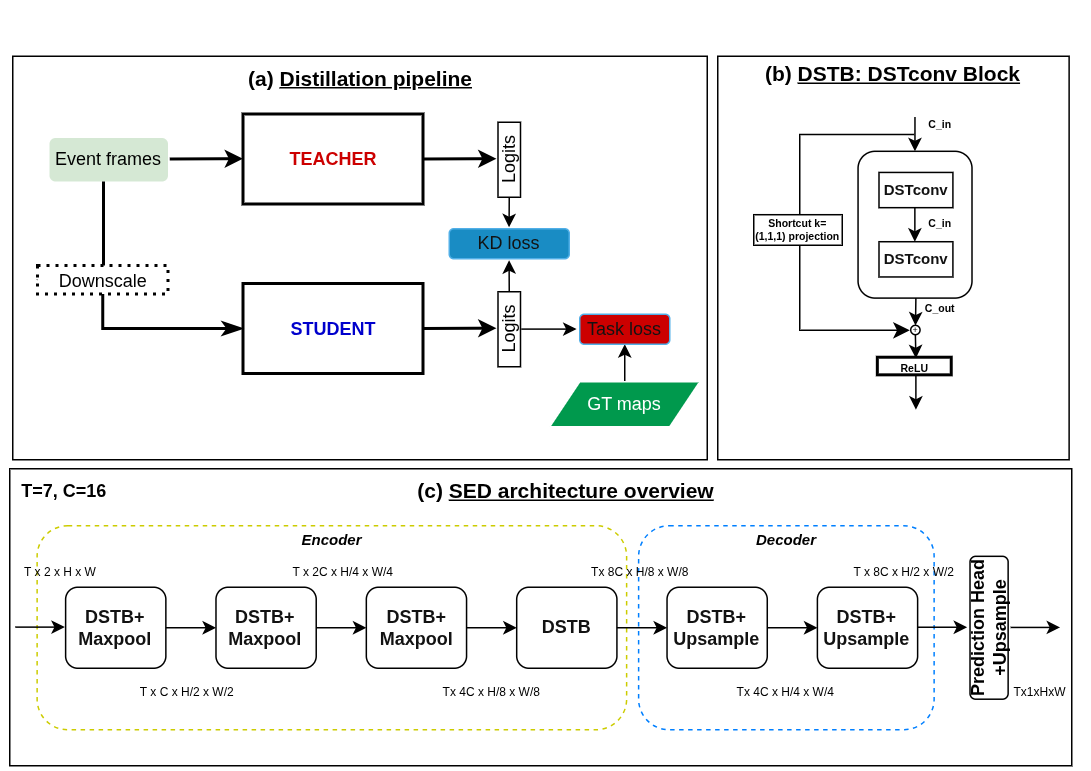}
    \caption{Overview: (a) Distillation pipeline, (b) DSTB consisting of two DSTconv modSED, and (c) SED architecture overview.}
    \label{fig:distill}
\end{figure*}
In this section, we present the SED architecture for efficient event-based saliency prediction. The model learns robust cross-domain representations through an explicit and efficient exploitation of the temporal information in event data, combined with knowledge distillation from a strong teacher. We design a simple yet efficient architecture combining depthwise-separable 3D convolutions and factorized spatio-temporal convolutions, exploiting temporal information at an
extremely reduced model size. To help the model stay robust under shifts to a new dataset or domain, we distill knowledge from SEST\cite{mazna2026exploringdeeplearningeventbased} leveraging the rich features of a pretrained transformer and decoding them into saliency maps. %We first describe our network, then our knowledge distillation scheme.

\subsection{Event representation}
Similar to~\cite{mazna2026exploringdeeplearningeventbased}, we construct a voxel grid representation from the raw events: a space-time histogram in which each voxel accumulates events falling within a given pixel and time interval. Given a stream of events $e_k = (x_k, y_k, t_k, p_k)$, where $(x_k, y_k)$ is the pixel location, $t_k$ the timestamp, and $p_k \in \{-1, +1\}$ the polarity, we divide the temporal window $[t_0, t_0 + \Delta t]$ into $T$ equal bins and assign each event to the
bin corresponding to its timestamp. The voxel grid $V \in \mathbb{R}^{T \times 2 \times H \times W}$ is formed by accumulating, for each polarity channel, the events that fall into each spatio-temporal cell:
\begin{equation}
V(\tau, p, x, y) \;=\;
\sum_{k} \mathbf{1}\!\left[\,p_k = p\,\right]\,
\mathbf{1}\!\left[\,x_k = x,\; y_k = y\,\right]\,
\mathbf{1}\!\left[\,b(t_k) = \tau\,\right],
\end{equation}
where $b(t_k) = \left\lfloor T \,\frac{t_k - t_0}{\Delta t} \right\rfloor$ maps an event timestamp to its temporal bin index $\tau \in \{0,\dots,T-1\}$, and the two polarity channels separate positive and negative brightness changes.

\subsection{Depthwise Spatio-Temporal Convolution}
\begin{figure*}[t]
    \centering
    \includegraphics[width=0.5\textwidth,
        keepaspectratio]{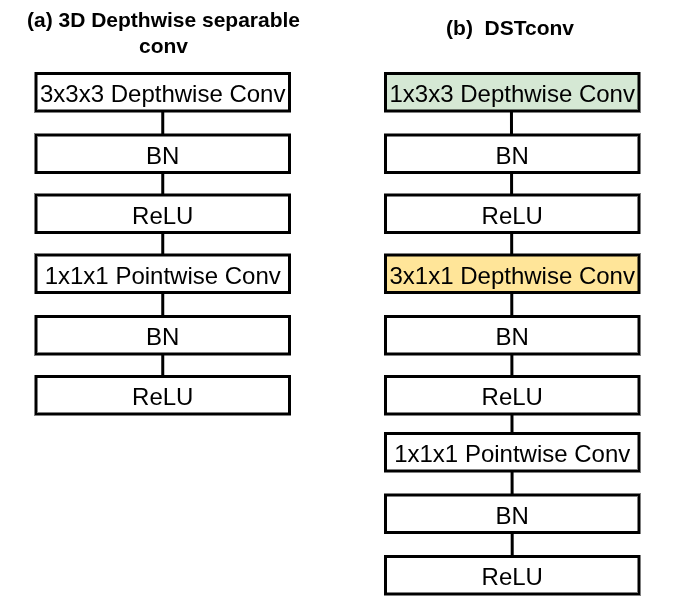}
    \caption{Detailed convolutional modSED. (a) 3D Depthwise separable conv, (b) DSTconv}
    \label{fig:DSTconv}
\end{figure*}

Existing compact models in the RGB domain are built mostly on MobileNets
\cite{howard2017mobilenets} and depthwise-separable convolutions
\cite{chollet2017xception}. In contrast, we want to explicitly exploit the
temporal information of event data, which prior work
\cite{mazna2026exploringdeeplearningeventbased} has shown to be important
for dynamic saliency prediction. A naive way to do so would be to use 3D
depthwise-separable convolutions \cite{ye20193d}; while these capture
temporal structure, their depthwise stage is $k$ times more expensive than
its 2D counterpart, where $k$ is the temporal kernel size. Driven by the
twin goals of explicit temporal modelling and extreme efficiency, and
inspired by~\cite{tran2018closer,lee2021diverse}, we factorize the
depthwise convolution of a 3D depthwise-separable convolution into separate
spatial and temporal depthwise convolutions. We call the resulting unit the
Depthwise Spatio-Temporal convolution (DSTconv), illustrated in Figure~\ref{fig:DSTconv}.

Let $F_{in} \in \mathbb{R}^{C \times T \times H \times W}$ be the input
feature. The DSTconv computes its output as
\begin{equation}
F_{out} = \mathrm{PW}\bigl(DW_t(DW_s(F_{in}))\bigr),
\end{equation}
where $DW_s$ is a spatial depthwise convolution with kernel
$1\times k_s \times k_s$, $DW_t$ is a temporal depthwise convolution with
kernel $k_t \times 1 \times 1$, and $\mathrm{PW}$ is a $1\times1\times1$
pointwise convolution that mixes channels. Each convolution is followed by
batch normalisation and a ReLU activation.

Our DSTconv is most closely related to the module of~\cite{lee2021diverse},
which also factorizes a depthwise convolution into spatial and temporal
stages. It differs in two ways: rather than applying a pointwise
convolution at the beginning of the block, we delay the single pointwise
operation to the end, reducing computation since pointwise mixing is the
most parameter-heavy operation; and our block uses no residual connection.
These choices reduce the parameter count drastically compared to other
factorization techniques while still exploiting the temporal information of
event data, allowing the network to easily meet the computational
requirements of the edge devices where event cameras excel. Finally, the
DSTconv reduces to a standard 2D depthwise-separable convolution when the
temporal depthwise stage $DW_t$ is removed.

For an input with $C$ channels and a layer that maps $C \to C'$ channels
with spatial kernel $k_s$ and temporal kernel $k_t$, the parameter counts of
the alternatives are:
\begin{align}
\text{Conv3D:} \quad      & C \cdot C' \cdot k_t k_s^2, \\
\text{3D DSConv:} \quad   & C \cdot k_t k_s^2 \;+\; C \cdot C', \\
\text{DSTconv (ours):} \quad & C \cdot k_s^2 \;+\; C \cdot k_t \;+\; C \cdot C'.
\end{align}
The factorization replaces the depthwise term $C k_t k_s^2$ with the much
smaller $C(k_s^2 + k_t)$, while the pointwise term $C C'$ -- which dominates
the count -- is used only once. %For our default $k_s = 3$, $k_t = 3$, this reduces the depthwise cost from $27C$ to $12C$ parameters per layer, a $2.25\times$ reduction in the depthwise stage on top of the savings already provided by depthwise separation.

\subsection{Lightweight network}
We design a model with four encoding stages and two decoding stages, shown
in Figure~\ref{fig:distill} (c). It follows an encoder--decoder structure with a
saliency prediction head. The decoder is smaller than the encoder, as the
literature has observed that a large decoder does not necessarily improve
accuracy, given that saliency maps carry less information than the input
frames~\cite{hu2020fastsal}. Each encoding stage contains a Depthwise Spatio-Temporal Block (DSTB) consisting of two DSTconv blocks
with a residual connection (see Figure \ref{fig:distill} (b)), and the number of feature channels is doubled
after each stage. The first three encoding stages are each followed by a
max-pooling operator that halves the spatial resolution. Each decoding stage
reduces the number of feature channels and applies a trilinear upsampling
operation that gradually restores the spatial resolution. Finally, a
$1\times1\times1$ Conv3D prediction head takes the output of the second
decoding stage and produces the saliency maps, which are smoothed with a
Gaussian blur and upsampled to the original shape.

\subsection{Knowledge distillation}
\label{sec:loss}
We distill knowledge from the teacher SEST into our student SED (see Figure \ref{fig:distill} (a). SEST leverages a pretrained swin transformer as backbone, extracting rich hierarchical features from the 4 four stages of the backbone and projecting them through a conv3d, and build on top of it a lightweight decoder consisting of two successive Conv3d to refine the features and predcits the saliency maps. SEST reported consistent and substantial improvements over prior methods. Our main
motivation for using knowledge distillation is to obtain a small model that
is also robust in terms of generalization -- important for event-based
saliency, where the ultimate goal is deployment on resource-limited devices
facing uncontrolled real-world scenarios.

%We initially experimented with distillation using both a target loss and intermediate-layer matching, but preliminary results showed that the feature-matching term contributed a very small loss that slightly degraded performance. We therefore retain only a target loss between the teacher and student logits.

Let $\mathcal{T}$ denote the frozen teacher SEST and $\mathcal{S}$ the
student. The teacher receives an input
$X_{\mathcal{T}} \in \mathbb{R}^{B \times T \times 2 \times 224 \times 224}$,
while the student receives a lower-resolution input
$X_{\mathcal{S}} \in \mathbb{R}^{B \times T \times 2 \times 128 \times 128}$,
where $B$ is the batch size, $T = 7$ the number of temporal bins, and the
third dimension is the event polarity. We train the student with a two-term
objective combining a task loss on the per-frame predictions and an
output-level distillation loss:
\begin{equation}
\mathcal{L} \;=\; \mathcal{L}_{\text{task}}
\;+\; \mathcal{L}_{\text{kd}}.
\label{eq:total}
\end{equation}
Let $\hat{y}_{\mathcal{S}} \in \mathbb{R}^{T \times H \times W}$ denote the
student's per-frame logits and
$\hat{y}_{\mathcal{T}} \in \mathbb{R}^{T \times H \times W}$ the teacher's
logits, and let $y \in \mathbb{R}^{T \times H \times W}$ denote the
ground-truth saliency maps.

The task term is the same objective used to supervise the teacher \cite{mazna2026exploringdeeplearningeventbased}. It
combines binary cross-entropy with the standard saliency metrics CC and
KLDiv~\cite{bylinskii2019metrics}, computed on the per-frame predictions:
\begin{equation}
\mathcal{L}_{\text{task}} \;=\;
\text{KL}(y\,\|\,\hat{y}_{\mathcal{S}})
\;-\; \alpha\,\text{CC}(\hat{y}_{\mathcal{S}}, y)
\;+\; \beta\,\text{BCE}(\hat{y}_{\mathcal{S}}, y),
\end{equation}
where $\alpha = 0.5$ and $\beta = 0.7$.

The output-level distillation loss matches the student's per-frame
predictions to the teacher's predictions:
\begin{equation}
\mathcal{L}_{\text{kd}} \;=\; \text{KL}(\hat{y}_{\mathcal{T}}\,\|\,\hat{y}_{\mathcal{S}}).
\end{equation}

%Unless otherwise specified, we set $\lambda_{\text{task}} = 1.0$ and $\lambda_{\text{kd}} = 1.0$.

\section{Experiments}
\label{sec:exps}

\subsection{Datasets and evaluation metrics}
We evaluate SED on the N-DHF1K \cite{mazna2026exploringdeeplearningeventbased} and N-UCF Sports \cite{mazna2026exploringdeeplearningeventbased} datasets, both simulated from RGB videos. N-DHF1K contains events generated from DHF1K, the largest video saliency dataset, spanning diverse video categories, while N-UCF Sports contains events from the UCF Sports
saliency dataset. N-DHF1K consists of 500 training, 100 validation, and 100 test videos; N-UCF Sports consists of 103 training, 15 validation, and 32 test videos. We quantitatively compare our model against state-of-the-art methods -- evST \cite{chane2024event}, SNNevProto \cite{dangeloEventDrivenBioinspired2022}, and SEST (the teacher within our framework) \cite{mazna2026exploringdeeplearningeventbased}.
Following the SEST paper, we also include five state-of-the-art RGB models to give a sense of the gap between the RGB and event domains.
Following standard practice~\cite{bylinskii2019metrics}, we adopt four widely used evaluation metrics. They comprise two location-based metrics -- Area Under the ROC Curve (Judd variant, AUC-J) and Normalized Scanpath Saliency (NSS) -- and two distribution-based metrics -- Pearson Correlation Coefficient (CC) and Similarity (SIM).

\subsection{Implementation details}
Our model was implemented and trained in PyTorch on a single NVIDIA H100 GPU. During distillation the teacher is frozen, while the student is trained with a learning rate of $0.02$, a batch size of $20$, and the AdamW optimizer. The teacher was pretrained on several event-frame counts ($7$, $10$, $14$, $21$); for efficiency we train only in the $7$-bin setup, so the teacher receives and predicts $7$ bins and the student receives $7$ bins and learns to predict $7$ maps from both the teacher's predictions and the ground-truth maps. Following~\cite{mazna2026exploringdeeplearningeventbased}, we generate bins at the original video sampling frequency, giving a bin duration of $100$\,ms for N-UCF Sports ($10$\,Hz) and $33.33$\,ms for N-DHF1K ($30$\,Hz). %While the teacher was pretrained on N-DHF1K, the student is distilled only on N-UCF Sports, to demonstrate the cross-dataset strength.

\begin{figure*}[t]
    \centering
    \includegraphics[width=\textwidth, height=0.6\textheight,
        keepaspectratio]{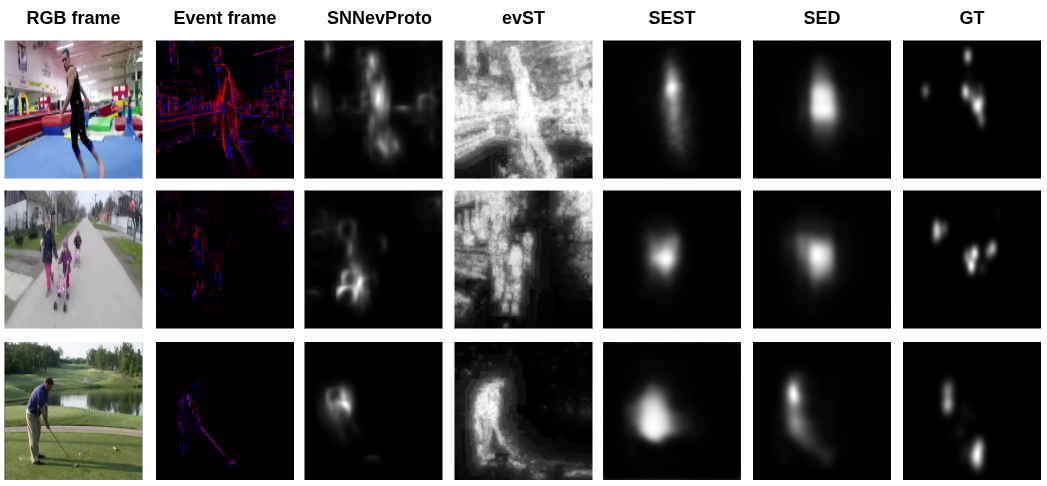}
    \caption{Qualitative results for three samples: First two rows (N-DHF1K), last row (N-UCF Sports). }
    \label{fig:viz}
\end{figure*}

\section{Results}
\label{sec:results}

\subsection{Comparison with state-of-the-art models}
Table~\ref{tab:main_comparison} reports the saliency metrics, model size, and
parameter count of SED against state-of-the-art event-based methods on
N-DHF1K and N-UCF Sports. We also include RGB-domain results to indicate the
remaining gap between domains; these are not directly comparable, as the test
splits and evaluation protocols differ from the event setting.

The first observation is that SED surpasses the teacher on all four saliency
metrics on N-UCF Sports, and on three of the four on N-DHF1K (CC, SIM, and
NSS). Only on AUC-J for N-DHF1K does the teacher remain ahead. It also
outperforms the other event-based methods across the board. This holds
despite SED's small size and memory footprint. On N-UCF Sports, the average
improvement over the teacher is roughly 5\% across metrics. These results not
only confirm prior findings that compact models can match the performance of
larger ones, but go further: in the event-based saliency setting, a compact
model can outperform a much larger teacher while drastically reducing
computational overhead

Figure~\ref{fig:viz} shows qualitative comparisons of SED against
state-of-the-art event-based saliency models. SED aligns well with the
ground truth and captures the semantic context of the scene, which is
important for real-time applications.

We further compare SED against its teacher SEST in terms of computational
efficiency (Table~\ref{tab:efficiency}), reporting parameters, model size,
MACs, and latency on both GPU and CPU. SED uses only $81$k parameters
($554\times$ fewer than SEST's $45$M), a $0.32$\,MB model size ($562\times$
smaller than $180$\,MB), and $353$M MACs ($447\times$ fewer), at an input
resolution of $128\times128$. To the best
of our knowledge, SED is the most compact model, in terms of parameter
count, for saliency prediction in both the RGB and event-based domains. We
measure latency with TensorRT at FP32. Under these conditions SED is
substantially faster than its teacher: $5.84$\,ms versus $32.9$\,ms on GPU,
and $38.89$\,ms versus $1175.6$\,ms on CPU.

\begin{table*}[htbp]
    \centering
    \caption{ Performance comparison of state-of-the-art models on N-UCF Sports and N-DHF1K. Results for models marked by $^\dagger$, $^\ddagger$, $^*$ are taken from \cite{moradi2026knowledge}, \cite{hu2023tinyhd}, \cite{mazna2026exploringdeeplearningeventbased} respectively.}
    \label{tab:main_comparison}
    \renewcommand{\arraystretch}{1.15}
    \setlength{\tabcolsep}{5pt}
    \resizebox{\textwidth}{!}{%
    \begin{tabular}{l|c|cccc|cccc|c|c}
        \toprule
        \multirow{2}{*}{\textbf{Method}} & \multirow{2}{*}{\textbf{Domain}} & \multicolumn{4}{c|}{\textbf{N-UCF Sports}} & \multicolumn{4}{c|}{\textbf{N-DHF1K}}   & \multirow{2}{*}{\textbf{Size (MB)}} & \multirow{2}{*}{\textbf{Params.}}\\
        & & AUC-J $\uparrow$ & CC $\uparrow$ & SIM $\uparrow$ & NSS $\uparrow$ & AUC-J $\uparrow$ & CC $\uparrow$ & SIM $\uparrow$ & NSS $\uparrow$ \\
        \midrule
        % --- RGB reference methods (dense frames; not directly comparable) ---
        \textit{OFF-ViNet}~\cite{ikenoya2024off} $^\dagger$ & RGB & 0.936 & 0.730 & 0.589 & 4.180 & 0.914 & 0.538 & 0.419 & 3.089 & -- & --\\
        \textit{Luo (2025)}~\cite{luo2025combining} $^\dagger$ & RGB & 0.939 & 0.720 & 0.570 & 3.988 & 0.913 & 0.526 & 0.388 & 3.008 & -- & 75M\\
        \textit{THTD-Net}~\cite{moradi2026knowledge} $^\dagger$ & RGB & 0.923 & 0.671 & 0.510 & 3.306 & 0.919 & 0.560 & 0.396 & 3.246 & 220 & --\\
        \textit{TinyHD}~\cite{hu2023tinyhd} $^\ddagger$ & RGB & 0.911 & 0.609 & 0.499 & 3.234 & 0.905 & 0.493 & 0.387 & 2.819 & -- & 3.92M \\
        \midrule
        % --- Event-based methods ---
        SNNevProto~\cite{dangeloEventDrivenBioinspired2022}$^*$ & Event & 0.8316 & 0.2834 & 0.1262 & 1.3286 & 0.5884 & 0.0375 & 0.1342 & 0.1725 & -- &-- \\
        evST~\cite{chane2024event} $^*$           & Event & 0.7787 & 0.2559 & 0.1263 & 1.3798 & 0.5380 & 0.0217 & 0.1317 & 0.0898 & -- & --\\
        SEST (teacher) ~\cite{mazna2026exploringdeeplearningeventbased} $^*$ & Event & 0.9094 & 0.5144 & 0.4280 & 2.7672 & \textbf{0.9197} & 0.4661 & 0.3634 & 2.3956 & 180 & 45M \\
        \midrule
        SED(ours) & Event & \textbf{0.9187} & \textbf{0.5703} & \textbf{0.4462} & \textbf{3.1418} & 0.9047 & \textbf{0.5618} & \textbf{0.4316} & \textbf{2.4667} & \textbf{0.32} & \textbf{81k}\\
        \bottomrule
    \end{tabular}%
    }
\end{table*}

\begin{table}[htbp]
    \centering
    \caption{Efficiency comparison between teacher and student models}
    \label{tab:efficiency}
    \renewcommand{\arraystretch}{1.15}
    \setlength{\tabcolsep}{5pt}
    \resizebox{\columnwidth}{!}{%
    \begin{tabular}{l|ccc|cc}
        \toprule
        \multirow{2}{*}{\textbf{Method}} & \multirow{2}{*}{\textbf{Params}} & \multirow{2}{*}{\textbf{MACs}} & \multirow{2}{*}{\textbf{Mem. (MB)}} & \multicolumn{2}{c}{\textbf{Latency (ms)}}\\ %& \multirow{2}{*}{\textbf{Reduction}} \\
        & & & & GPU & CPU \\
        \midrule
        SEST (teacher)    & 45M & 158G   & 180.0 & 32.9 & 1175.6 \\% & 1$\times$ \\
        SED(ours) & \textbf{81k} & \textbf{353M} & \textbf{0.32} & \textbf{5.84} & \textbf{38.89}\\% & \textbf{562$\times$} \\
        \bottomrule
    \end{tabular}%
    }
\end{table}

\subsection{Effect of knowledge distillation}
We now analyze how much knowledge distillation contributes, by comparing
each distilled model against an identical one trained from scratch
(Table~\ref{tab:kd_generalization}). We first observe that the scratch
model learns well and fits the dataset it is trained on. On N-UCF Sports,
however, the distilled model outperforms the scratch model on all metrics
except AUC-J. On N-DHF1K, the two are comparable in-domain. This shows that
a small model can learn on its own, but only the specific distribution it
was trained on -- reinforcing the value of distillation in our setting.

We then assess distillation beyond the training distribution. Distillation
strongly improves performance on each dataset when the model has not been
trained on it. Most notably, the model distilled on N-UCF Sports -- a small,
narrow dataset -- transfers strongly to N-DHF1K, which is larger and more
diverse and which the student never saw directly (only the teacher was
trained on it). Trained from scratch on N-UCF Sports, the student collapses when evaluated
on N-DHF1K: NSS drops sharply from $3.05$ in-domain to $0.90$, CC from $0.53$
to $0.22$, and SIM from $0.44$ to $0.24$. The distilled model, by contrast,
retains an NSS of $2.01$, CC of $0.47$, and SIM of $0.37$. This suggests that
distillation lets the student acquire more than the dataset-specific patterns
it would learn from the ground truth alone.

Finally, we evaluate generalization to the real event dataset EBSD
\cite{chane2024event} (Table~\ref{tab:kd_generalization}). Despite the domain shift from
synthetic to real events, the distilled model generalizes well -- exceeding
even the teacher on all reported metrics -- while the scratch model fails to
transfer. Notably, distillation on N-UCF Sports alone remains competitive
with distillation on the larger N-DHF1K, confirming that the gains come from
the distillation process rather than from the scale of the training set.
\begin{table*}[htbp]
    \centering
    \caption{Effect of knowledge distillation on cross-dataset generalization.}
    \label{tab:kd_generalization}
    \renewcommand{\arraystretch}{1.2}
    \setlength{\tabcolsep}{5pt}
    \resizebox{\textwidth}{!}{%
    \begin{tabular}{ll|cccc|cccc|cccc}
        \toprule
        \multirow{2}{*}{\textbf{Train data}} & \multirow{2}{*}{\textbf{Regime}}
        & \multicolumn{4}{c|}{\textbf{N-UCF Sports}}
        & \multicolumn{4}{c|}{\textbf{N-DHF1K}}
        & \multicolumn{4}{c}{\textbf{EBSD (real)}} \\
        & & AUC-J & CC & SIM & NSS & AUC-J & CC & SIM & NSS & AUC-J & CC & SIM & NSS \\
        \midrule
        \multirow{1}{*}{N-DHF1K}
        & SEST (Scratch)     & 0.8841 & 0.4555 & 0.3214 & 2.4537 & 0.9197 & 0.4661 & 0.3634 & 2.3956 & 0.8518 & 0.5304 & 0.4922 & 1.4456 \\
        \midrule
        \multirow{2}{*}{N-UCF Sports}
          & Scratch     & 0.9224 & 0.5266 & 0.4390 & 3.0526 & 0.8129 & 0.2184 & 0.2397 & 0.8981 & 0.7928 & 0.2947 & 0.3708 & 0.7478 \\
          & Distilled   & 0.9187 & 0.5703 & 0.4462 & 3.1418 & 0.8900 & 0.4697 & 0.3683 & 2.0121 & 0.8568 & 0.5443 & 0.5123 & 1.4531 \\
        \midrule
        \multirow{2}{*}{N-DHF1K}
          & Scratch     & 0.8627 & 0.4031 & 0.3388 & 2.0018 & 0.9043 & 0.5665 & 0.4291 & 2.4814 & 0.8758 & 0.6787 & 0.5840 & 1.9309 \\
          & Distilled   & 0.8915 & 0.4778 & 0.3928 & 2.4450 & 0.9047 & 0.5618 & 0.4316 & 2.4616 & 0.8817 & 0.6746 & 0.5828 & 1.8707 \\
        \bottomrule
    \end{tabular}%
    }
\end{table*}

\subsection{Ablation studies}
To experimentally validate our design choices, we carry out ablation studies
on SED: replacing the DSTconv block with a depthwise-separable 2D convolution
(DS2D) and with a depthwise-separable 3D convolution (DS3D), and varying the
resolution of the input event frames.

\paragraph{DSTconv block.}
Here we replace DSTconv with DS2D and DS3D while keeping the rest of the model
configuration fixed. For DS2D, the temporal dimension $T$ is folded into the
channel dimension and recovered at the end of the block.
Table~\ref{tab:block_ablation} shows that on N-UCF Sports, DSTconv outperforms
DS2D, while DS3D performs best but at a substantially higher computational
cost ($2.63$\,G MACs versus $353$\,M for DSTconv). On N-DHF1K, DSTconv and DS2D
both outperform DS3D: DS2D is slightly ahead on AUC-J and SIM, while DSTconv
leads on CC and NSS. Overall, DSTconv matches or exceeds DS2D in accuracy while
remaining far cheaper than DS3D, offering the best accuracy--cost trade-off.

\begin{table*}[htbp]
    \centering
    \caption{Ablation of the convolutional block. We compare a 2D depthwise-separable block (DS2D), a 3D depthwise-separable block (DS3D), and our Depthwise Spatio-Temporal Block (DSTconv).}
    \label{tab:block_ablation}
    \renewcommand{\arraystretch}{1.15}
    \setlength{\tabcolsep}{5pt}
    \resizebox{\textwidth}{!}{%
    \begin{tabular}{l|ccc|cccc|cccc}
        \toprule
        \multirow{2}{*}{\textbf{Block}} & \multirow{2}{*}{\textbf{Params}} & \multirow{2}{*}{\textbf{MACs}} & \multirow{2}{*}{\textbf{Mem. (MB)}} & \multicolumn{4}{c|}{\textbf{N-UCF Sports}} & \multicolumn{4}{c}{\textbf{N-DHF1K}} \\
        & & & & AUC-J $\uparrow$ & CC $\uparrow$ & SIM $\uparrow$ & NSS $\uparrow$ & AUC-J $\uparrow$ & CC $\uparrow$ & SIM $\uparrow$ & NSS $\uparrow$ \\
        \midrule
        
        DS2D        & 79k & 14M & 0.32 & 0.9233 & 0.5531 & 0.4451 & 3.0591 & 0.9051 & 0.5605 & 0.4324 & 2.4569 \\
        
        DS3D        & 90k & 2.63G & 0.36 & 0.9340 & 0.5905 & 0.4644 & 3.3346 & 0.9040 & 0.5529 & 0.4303 & 2.4153 \\
        
        DSTconv (ours) & 81k & 353M & 0.32 & 0.9187 & 0.5703 & 0.4462 & 3.1418 & 0.9047 & 0.5618 & 0.4316 & 2.4667 \\
        \bottomrule
    \end{tabular}%
    }
\end{table*}

\paragraph{Input resolution.}
We evaluate the effect of input event-frame resolution on N-UCF Sports,
feeding the model smaller ($64\times64$) and larger ($224\times224$) frames
in addition to our default ($128\times128$). Table~\ref{tab:input_resolution}
shows that $128\times128$ gives the best overall accuracy. Interestingly, the larger $224\times224$ input degrades performance more than
the smaller $64\times64$ one. We hypothesize that because event data is
spatially sparse, a higher resolution spreads the active events across more
pixels and increases the proportion of empty (zero) pixels, weakening the
signal, whereas a moderate resolution keeps the events spatially denser. This makes $128\times128$ a favorable
operating point in both accuracy and compute

\begin{table}[htbp]
    \centering
    \caption{Effect of student input resolution on the N-UCF Sports.}
    \label{tab:input_resolution}
    \renewcommand{\arraystretch}{1.2}
    \setlength{\tabcolsep}{6pt}
    \begin{tabular}{l|cc|cccc}
        \toprule
        \textbf{Input res.} & \textbf{MMACs} & \textbf{Latency (ms)} & AUC-J $\uparrow$ & CC $\uparrow$ & SIM $\uparrow$ & NSS $\uparrow$ \\
        \midrule
        $64\times64$   & 88 & 2.51 & 0.9274 & 0.5685 & 0.4584 & 2.9983 \\
        $128\times128$ & 353  & 5.84 & 0.9187 & 0.5703 & 0.4462 & 3.1418  \\
        $224\times224$ & 2160 & 35.79 & 0.9168 & 0.5110 & 0.3744 & 2.7792 \\
        \bottomrule
    \end{tabular}
\end{table}

\section{Conclusion}
\label{sec:concl}

In this work, we introduced SED, a lightweight network for event-based
saliency prediction trained via knowledge distillation. With a model size of
only $0.32$\,MB -- a $562\times$ reduction relative to its teacher -- SED
achieves state-of-the-art performance on the N-DHF1K and N-UCF Sports
datasets. The model builds on an efficient spatio-temporal block that reduces
the parameter count by $554\times$ and the MACs by $447\times$ relative to the teacher, reaching $81$k
parameters and $353$M MACs at an input resolution of $128\times128$. Beyond
efficiency, SED generalizes strongly to unseen data: distilled on the small
N-UCF Sports set alone, it transfers to the larger N-DHF1K and to real event
recordings, where an identical model trained from scratch collapses. These
properties make it well suited to resource-constrained applications. Future
work will focus on hardware deployment for real-time on-device inference. We
hope this work encourages the development of more efficient event-based
perception models and, in turn, more efficient edge systems.

\bibliographystyle{plain}
\bibliography{main, refs}
\end{document}